\def\year{2019}\relax
\documentclass[letterpaper]{article} %DO NOT CHANGE THIS
\usepackage{aaai19}  %Required
\usepackage{times}  %Required
\usepackage{helvet}  %Required
\usepackage{courier}  %Required
\usepackage{url}  %Required
\usepackage{graphicx}  %Required
\usepackage{url}
\usepackage{multirow}
\usepackage{graphicx}
\usepackage{hyperref}

\newcommand{\beginsupplement}{%
        \setcounter{table}{0}
        \renewcommand{\thetable}{A\arabic{table}}%
        \setcounter{figure}{0}
        \renewcommand{\thefigure}{A\arabic{figure}}%
        \setcounter{section}{0}
        \renewcommand{\thesection}{A\arabic{section}}%
     }

\newcommand{\citet}[1]{\citeauthor{#1}~\shortcite{#1}}
\newcommand{\citep}{\cite}

\begin{document}
\title{What If We Simply Swap the Two Text Fragments?\\A Straightforward yet Effective Way to Test the Robustness of Methods to Confounding Signals in Nature Language Inference Tasks}
% \author{AAAI Press\\
% Association for the Advancement of Artificial Intelligence\\
% 2275 East Bayshore Road, Suite 160\\
% Palo Alto, California 94303\\
% }
\author{Haohan Wang$^1$, Da Sun$^2$, Eric P. Xing$^{1,3}$\\
$^1$Language Technologies Institute, School of Computer Science, Carnegie Mellon University,\\ Pittsburgh, PA, USA\\
$^2$School of Information Science, Southeast University\\
Nanjing, China \\
$^3$Machine Learning Department, School of Computer Science, Carnegie Mellon University,\\ Pittsburgh, PA, USA\\
\textit{haohanw@cs.cmu.edu}
}
\maketitle
\begin{abstract}
Nature language inference (NLI) task is a predictive task of determining the inference relationship of a pair of natural language sentences. With the increasing popularity of NLI, many state-of-the-art predictive models have been proposed with impressive performances. However, several works have noticed the statistical irregularities in the collected NLI data set that may result in an over-estimated performance of these models and proposed remedies. In this paper, we further investigate the statistical irregularities, what we refer as confounding factors, of the NLI data sets. With the belief that some NLI labels should preserve under swapping operations, we propose a simple yet effective way (swapping the two text fragments) of evaluating the NLI predictive models that naturally mitigate the observed problems. Further, we continue to train the predictive models with our swapping manner and propose to use the deviation of the model's evaluation performances under different percentages of training text fragments to be swapped to describe the robustness of a predictive model. Our evaluation metrics leads to some interesting understandings of recent published NLI methods. Finally, we also apply the swapping operation on NLI models to see the effectiveness of this straightforward method in mitigating the confounding factor problems in training generic sentence embeddings for other NLP transfer tasks. 
\end{abstract}

\noindent Natural Language Inference (NLI) task is testing the ability of a computational model to understand the natural language by the evaluation of a three-way classification for two fragments of natural language texts \citep{bowman2015large}. The two fragments, namely \textit{premise} (P) and \textit{hypothesis} (H), are labeled in three different categories (\textit{i.e.} \textit{entailment} (E), \textit{neutral} (N), and \textit{contradiction} (C)) based on their entailment relation. More specifically, we can have:
\begin{itemize}
    \item $E$: $H$ is definitely true given $P$
    \item $C$: $H$ is definitely not true given $P$
    \item $N$: $H$ may or may not be true given $P$
\end{itemize}
Formally, we could rewrite the natural language described linguistic entailment relation into propositional logic as following:
\begin{itemize}
\item $E$: $P$ $\rightarrow$ $H$
\item $C$: $P$ $\rightarrow$ $\neg$ $H$
\item $N$: $P$ $\perp$ $H$
\end{itemize}
where $\rightarrow$ stands for implication, $\neg$ stands for negation, and we use $\perp$ to denote that there is no clear relation between $P$ and $H$. 

The essential part of this paper lies in the fact that for any two propositions $A$ and $B$, we have:
\begin{itemize}
    \item ($A$ $\rightarrow$ $B$) $\perp$ ($B$ $\rightarrow$ $A$)
    \item ($A$ $\rightarrow$ $\neg$ $B$) $\iff$ ($B$ $\rightarrow$ $\neg$ $A$)
    \item ($A$ $\perp$ $B$) $\iff$ ($B$ $\perp$ $A$)
\end{itemize}
In simpler words, swapping $A$ and $B$ will retain the $\rightarrow$ $\neg$ relation and $\perp$ relation, but not $\rightarrow$ relation. Therefore, we can simply evaluate an NLI predictive model by swapping the \textit{premise} and \textit{hypothesis} in testing data set with the argument: \textbf{If a model can truly predict inference relationship between pairs of text fragments, it should report comparable accuracy between the original test set and swapped test set for \textit{contradiction} pairs and \textit{neutral} pairs, and lower accuracy in swapped test set for \textit{entailment} pairs.} If we do not observe such an accuracy pattern, it is very likely that the evaluated NLI predictive model is questionable despite its impressive performance. One may argue that the label may not necessarily preserve for \textit{neutral} pairs after swapping, we explain in the Discussion section later.

This work is inspired by several recent works that have observed the statistical irregularities of constructed NLI data sets. For example, \citet{gururangan2018annotation} noticed that during the data construction phase, the workers create a heuristic to generate hypothesis with minimum risks of being wrong, which lead to the fact that there are different distributions of words for different labels, and a powerful model can easily bypass the semantic information of the texts and predict the label with reasonable accuracy. The similar phenomenon has been observed by \citet{poliak2018hypothesis}, who noticed that a model could infer the NLI label with only hypotheses, and proposed a hypothesis-only baseline. Recently, \citet{naik2018stress} also noticed that the machine learning models might exploit the idiosyncrasies in the construction of the data set in predicting NLI labels and proposed a ``stress test'' to evaluate whether the models can make semantic-level inferences. All these previous work have discussed the inherent limitations of current NLI tasks and proposed solutions such as more comprehensive evaluation criteria \citep{naik2018stress}, guidance in future constructions of data sets \citep{gururangan2018annotation}, or more powerful methods such as utilizing the propositional logic relationship to regularize the neural network to ignore statistical irregularities of words and predict through semantic-level information \citep{minervini2018adve}. While this paper agrees with the necessity of these carefully-designed remedies, we want to bring the community's attention to a straightforward, easily-implemented, yet meaningful solution: swapping the two text fragments.  

% In this paper, inspired by recently reported misalignments between the conceptual goal and statistical realization of many computational models \citep{wang2016select,goyal2017making,poliak2018towards}, we aim to further answer the question about the ability of a computational model to understand the natural language. We utilize the equivalence of propositional logic rules to propose a more carefully designed, yet simple to implement, evaluation metric to help evaluation whether a computational model truly understands the natural language text fragments. 

Our contribution of this paper can be summarized in three-fold:
\begin{itemize}
\item By recognizing the spurious signals between sentences' semantic and words' distribution of \textit{hypothesis} created through a confounding factor of word choices, we introduce a straightforward and effective way to break such dependency: swapping the two text fragments. We apply the swapping evaluation metric on several recent NLI models and notice that some methods seem to predict based on local statistical patterns, in contrast to semantic level understanding. 
\item Inspired by the swapping evaluation metric, we propose to train the NLI models with a sequence of different training data sets that are defined by the percentages of sentence pairs swapped in the training set. We examine the ``deviation'' of performances of the model trained by these data sets evaluated by more powerful metrics. We propose the ``deviation'' to be a good measure of the robustness of the models to confounding signals.  
\item We further investigate how the swapping training procedure help mitigate the confounding factor problem during NLI training by applying the trained sentence embeddings to other NLP transfer tasks.  
\end{itemize}

The remainder of the paper is organized as follows. We first introduce the background of this paper, where we explain why swapping the two text fragments is necessary, rather than just a random trick we adopt, and we also explain the concept of confounding factors in the context of NLI. Then we introduce the swap evaluation and the results, which further disclose some mechanism of the evaluated models. Further, we continue to test the robustness of the models to confounding signals. Afterward, we test to see if the swapping training in NLI will lead to more meaningful sentence embeddings that help in other NLP transfer tasks. Finally, with a brief discussion of the related work, we conclude the paper. 

% The rest of this paper is organized as following: we first discuss the related works that have observed the statistical irregularities in NLI task in Section~\ref{sec:related}. Then in Section~\ref{sec:eval} we start to introduce the SWAP evaluation in details. We also re-evaluate some of the state-of-the-art models to show that the good predictive performance neural networks achieve in NLI tasks are only partially due to the ability to learn the meaning of the sentence. In Section~\ref{sec:train}, inspired by the evaluation procedure, we introduce the SWAP training procedure that helps in improving the performance in NLI tasks in both conventional evaluation and our newly proposed evaluation metric. We further apply the training procedure to several different downstream tasks and achieve the superior performance in Section~\ref{sec:results}. Finally, the conclusion is drawn in Section~\ref{sec:con}. 

\begin{figure}
    \centering
    \includegraphics[width=1.0\columnwidth]{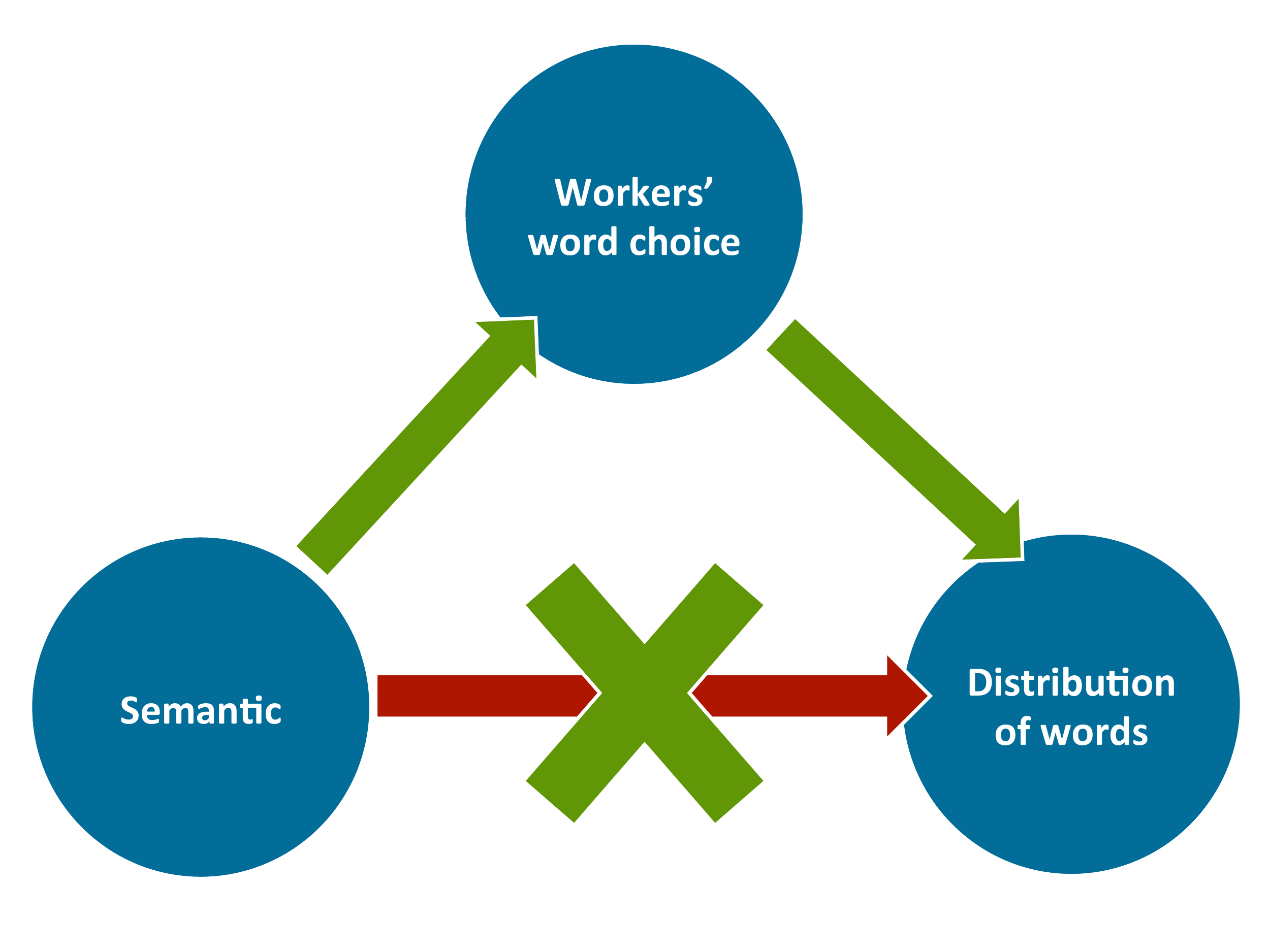}
    \caption{Illustration of the confounded relationship between the words' distribution and the semantic information: The Amazon Mechanical Turk workers choose which words to use according to the inference relationship they are asked, and their choice affects the distribution of the words. As a result, a spurious relationship is created between the semantic and words' distribution (denoted as the red arrow). }
    \label{fig:confounder}
\end{figure}

\section{Background: Artifacts in NLI and Confounding Factors}
\label{sec:back}
The NLI data set (\textit{e.g.} the SNLI data set introduced by \citet{bowman2015large}) is constructed with the help of Amazon Mechanical Turk workers by presenting the worker with one sentence (\textit{premise}) and asking the workers to write sentences (\textit{hypothesis}) that are:
\begin{itemize}
    \item definitely true given the presented sentence, for \textit{entailment},
    \item definitely not true given the presented sentence, for \textit{contradiction}
    \item might be true given the presented sentence, for \textit{neutral}
\end{itemize}

As \citet{gururangan2018annotation} pointed out, these workers seem to find convenient ways to write \textit{hypothesis}, such as using negation words (such as ``no'', ``nobody'', and ``nothing'') to highlight the \textit{contradiction} relation, or using generic words to replace specific words (such as ``animal'' for ``dog'', ``instrument'' for ``guitar'', and ``outdoors'' for ``beach'') to guarantee the \textit{entailment} relation. Therefore, these workers create different word distributions according to the different labels. %Also, as \citet{gururangan2018annotation} pointed out, simply removing the pairs with troublesome word choices will likely create other artifacts, and the most effective solution seems to be construct another data set by offering sufficient amount of guidance to the workers. 

As a result, there exist confounding factors in the NLI data set. As we can see in Figure~\ref{fig:confounder}, the semantic label leads the workers' choices of words, which further directly determines the word distributions of \textit{hypothesis}. Therefore, it creates a spurious signal that the distribution of words, in addition to the semantic meaning of these sentences, is also related to the NLI label. If the machine learning models capture this spurious relation, the models will report impressive performances if evaluated regularly with NLI testing data set but will exhibit less favorable performances if evaluated with sophisticated methods. 

By acknowledging these background information, one should notice that the swapping operation we propose in this paper is more than a simple data augmentation trick. It serves as an immediate solution for the aforementioned confounding factor problem: by swapping \textit{premise} and \textit{hypothesis} in test data set, we change the word distributions between training set \textit{hypothesis} and test set \textit{hypothesis}, therefore, models that can only predict through these spurious signals will unlikely be effective in the new test set. 

\begin{table*}[]
\centering
\caption{Swapping evaluation results on several NLI models. Swap-$\star$ denotes the swapped data sets. Diff-$\star$ denotes the performance drop of evaluated accuracy from the original data set to the swapped data set. We can observe the performance drop in many cases, some of which are significant. These results may indicate that these methods have been over-estimated in the ability in predicting the relationship of sentence pairs at the semantic level. }
\label{tab:result1}
\begin{tabular}{cccccccc}
\hline
Model & Label & Dev & Swap-Dev & Diff-Dev & Test & Swap-Test & Diff-Test \\ \hline
\multirow{3}{*}{CBOW} & E & 0.877 & 0.134 & 0.743 & 0.856 & 0.080 & 0.776 \\
 & C & 0.706 & 0.583 & 0.123 & 0.740 & 0.580 & 0.160 \\
 & N & 0.874 & 0.613 & 0.261 & 0.659 & 0.589 & 0.070 \\ \hline
\multirow{3}{*}{InferSent} & E & 0.850 & 0.090 & 0.760 & 0.880 & 0.087 & 0.793 \\
 & C & 0.853 & 0.666 & 0.187 & 0.859 & 0.682 & 0.177 \\
 & N & 0.795 & 0.713 & 0.082 & 0.795 & 0.712 & 0.083 \\ \hline
\multirow{3}{*}{DGA} & E & 0.822 & 0.376 & 0.446 & 0.854 & 0.422 & 0.432 \\
 & C & 0.720 & 0.660 & 0.060 & 0.711 & 0.650 & 0.061 \\
 & N & 0.700 & 0.648 & 0.052 & 0.700 & 0.619 & 0.081 \\ \hline
\multirow{3}{*}{ESIM} & E & 0.891 & 0.301 & 0.590 & 0.884 & 0.324 & 0.560 \\
 & C & 0.865 & 0.702 & 0.163 & 0.861 & 0.701 & 0.160 \\
 & N & 0.806 & 0.721 & 0.085 & 0.801 & 0.720 & 0.081 \\ \hline
\multirow{3}{*}{KIM} & E & 0.908 & 0.103 & 0.805 & 0.895 & 0.095 & 0.800 \\
 & C & 0.850 & 0.772 & 0.078 & 0.845 & 0.796 & 0.049 \\
 & N & 0.800 & 0.664 & 0.136 & 0.781 & 0.675 & 0.106 \\ \hline
\multirow{3}{*}{ADV} & E & 0.862 & 0.856 & 0.006 & 0.854 & 0.860 & -0.006 \\
 & C & 0.753 & 0.643 & 0.110 & 0.751 & 0.646 & 0.105 \\
 & N & 0.706 & 0.509 & 0.197 & 0.705 & 0.507 & 0.198 \\ \hline
\end{tabular}
\end{table*}

\section{Swapping Evaluation and Results}
\label{sec:eval}
We proceed to officially introduce the Swapping evaluation method, what we expect, the results we have tested on recent published state-of-the-art methods, and follow-up analyses. 

Despite the importance of correcting the artifacts, our method is as simple as swapping the \textit{premise} and \textit{hypothesis}. However, one should notice we do not always expect high scores in the evaluation. For a model that will predict the pairs on semantic levels, we expect:
\begin{itemize}
    \item (significant) drop of performance for \textit{entailment} pairs,
    \item roughly the same performance for \textit{contradiction} pairs,
    \item roughly the same performance for \textit{neutral} pairs. 
\end{itemize}
Also, it's worth pointing out that a model that does not introduce performance drop in \textit{entailment} pairs may not be used as an evidence to show the robustness of this model, but should be used as evidence to show that the model or the data may have some other interesting properties that we need further investigate. 

We applied this evaluation method onto the following six different methods:
\begin{itemize}
    \item CBOW: An MLP that uses averaged continuous bag of words as features.
    %\item BiLSTM: A classifier that uses averaged word representation extracted from a conventional LSTM as features.  
    \item InferSent\footnote{https://github.com/facebookresearch/InferSent} \citep{conneau2017supervised}: It consists of sentence embedding, sequence encoder, composition layer, and the top layer classifier. The top layer classifier is an MLP whose input is a concatenation of \textit{hypothesis} representation, \textit{premise} representation, the dot product of these two, and the absolute value of the difference of these two. 
    \item DGA (Deep Gated Attention Bidirectional LSTM)\footnote{https://github.com/lukecq1231/enc\_nli/} \citep{chen2017recurrent}: The structure is similar to InferSent. The composition layers involves an operation named Gated-attention, which is inspired by the fact that human tends to remember only parts of the sentence after reading. 
    \item ESIM (Enhanced Sequential Inference Model)\footnote{https://github.com/lukecq1231/nli} \citep{chen2016enhancing}: A method that introduces local inference modeling, which models the inference relationship between \textit{premise} and \textit{hypothesis} after the two fragments aligned locally. 
    \item KIM (Knowledge-based Inference Model)\footnote{https://github.com/lukecq1231/kim} \citep{chen2018neural}: This model enriches ESIM with external knowledge. At this moment, the external knowledge includes lexical semantic relation, including whether two words are synonymy, antonymy, hypernymy, hyponymy \textit{etc}. 
    \item ADV (Adversarially Regularized Neural NLI Model)\footnote{https://github.com/uclmr/adversarial-nli/blob/master/nnli/parser.py} \citep{minervini2018adve}: an NLI model that is trained to minimize the standard loss as well as an inconsistency loss, which measures the model's performance over the adversarial set of data generated following logical rules.  
\end{itemize}

The results\footnote{All the experiments are run with standard Amazon EC2 p2.xlarge servers with Deep Learning AMI Version 13.0. The experiments are run with the default parameter settings set in the corresponding GitHub repositories.} are reported in Table~\ref{tab:result1}. We can observe the performance drop in many cases, some of which are significant. Therefore, following our previous argument, the results, unfortunately, indicate the potential incompetence of some of these evaluated models in capturing the semantic-level information despite these models' impressive performances with the original evaluation metric. 

It is relieving to notice that for \textit{contradiction} pairs and \textit{neutral} pairs, the models can still predict with an accuracy that is better than random chances. We believe this is because that the model can still capture a certain amount of semantic information from data to achieve an above-random prediction accuracy. 

We notice that the performance drop of DGA is quite minor, which may indicate that the gated-attention mechanism can help exclude the signals from superficial statistical signals such as word distributions. 

Also, we notice that for the methods InferSent and ESIM, the performance drop of \textit{contradiction} pairs is greater than the performance drop of \textit{neutral} pairs, but this trend is reversed for KIM and ADV. KIM and ADV are relatively new methods, therefore, generally believed to be more powerful than other methods. Interestingly, KIM relies on external knowledge in lexical semantic relation, and ADV is regularized by extra adversarial data that are generated according to logical rules. There might exist two explanations for KIM and ADV: 1) We conjecture that the extra information used by KIM and ADV helps eliminate the statistical irregularities in \textit{contradiction} pairs, but introduce extra spurious signals for \textit{neutral} pairs; 2) it is possible that KIM and ADV can more reliably ignore spurious signals than other methods because, as we will explain in detail in the Discussion section, some \textit{neutral} pairs may not preserve the label after the swapping operation so certain amount of performance drop is expected. 

Interestingly, we notice that there is barely any performance drop for ADV in the \textit{entailment} pairs after swapping while all the other methods see a significant performance drop. As we argued previously, it might not be good if the performance does not drop when we expect so. This result indicates that we may need to investigate further into the mechanism of ADV.  

While more error analysis are presented in the Appendix. To conclude this section, we propose a straightforward swapping evaluation method for methods of the NLI task. With our evaluation method, we can quickly tell some interesting properties of the current NLI methods. 

\section{Robustness Test to Confounding Factors Through Stress Test}
\label{sec:robust}

Following previous sections where we showed that swapping evaluation can serve as an effective measure in determining whether a model predicts via spurious signals because it can break the dependency between sentence semantic and the words' distribution of \textit{hypothesis}, we continue to ask the question of whether the NLI models are robust to the confounding signals. The essential argument of this section is: \textbf{For an NLI model $M$, given a powerful metric $T$ that can evaluate the NLI methods at the semantic level ignoring signals of confounding factors, if $T$ reports similar performances of $M_i$, when $M$ is trained repeatedly with different training data sets defined by the different percentages ($i$) of text pairs swapped, we can conclude that this model is robust to the confounding signals of words' distribution.}

To explain this argument, if an NLI model $M$ is learning through confounding signals, when such confounding signals are mitigated by swapping $i$ percentage of the training pairs during training, we should observe changes of the evaluated performance when $M_i$ is evaluated by $T$ in comparison of $M_0$ evaluated by $T$. We define the overall magnitude of relative changes of $M_i$ for multiple $i$ as the robustness of $M$ to the confounding signals of word distribution. 

Also, note that a model is robust to confounding factors does not necessarily mean that the model can always predict via semantics. It only means the model is not focusing on spurious signals. It is possible that a model learns neither the semantic information nor the confounding information (\textit{e.g.} a model that always predicts randomly). 

Additionally, note that due to the label preserving property for different labels we discussed in the previous section, only the \textit{contradiction} and \textit{neutral} pairs can be swapped during training. 

\begin{table*}[]
\centering
\caption{``Stress Test'' results for different percentage of training text fragments to be swapped. S$\star$ denotes the percentage of text fragments are randomly swapped during training (Only \textit{neutral} and Contrdiction pairs can be swapped). Accuracies shown on both genre-matched and mismatched categories for each stress test. R$\star$ denotes the ratio of results over the case when no pairs are swapped (S0\%). Devi (last column) denotes the overall deviation of R$\star$ from 100\% within each test.}
\label{tab:result3}
\begin{tabular}{ccccccccccccc}
\hline
Model & Test & Cat & S0\% & S25\% & R25\% & S50\% & R50\% & S75\% & R75\% & S100\% & R100\% & Devi \\ \hline
\hline 
\multirow{10}{*}{InferSent} & \multirow{2}{*}{Antonymy} & Mat & 22.87 & 24.92 & 109\% & 26.20 & 115\% & 27.61 & 121\% & 24.47 & 107\% & \multirow{2}{*}{0.173} \\
 &  & Mis & 16.92 & 18.45 & 109\% & 19.61 & 116\% & 21.57 & 127\% & 17.82 & 105\% &  \\ \cline{2-13} 
 & Length & Mat & 56.95 & 58.08 & 102\% & 57.11 & 100\% & 57.91 & 102\% & 57.26 & 101\% & \multirow{2}{*}{0.000} \\
 & Mismatch & Mis & 58.31 & 58.84 & 101\% & 58.07 & 100\% & 58.94 & 101\% & 58.54 & 100\% &  \\ \cline{2-13} 
 & \multirow{2}{*}{Negation} & Mat & 48.78 & 49.60 & 102\% & 49.65 & 102\% & 49.45 & 101\% & 50.24 & 103\% & \multirow{2}{*}{0.002} \\
 &  & Mis & 48.09 & 49.08 & 102\% & 49.15 & 102\% & 48.40 & 101\% & 48.99 & 102\% &  \\ \cline{2-13} 
 & Word & Mat & 55.58 & 56.86 & 102\% & 56.96 & 102\% & 54.21 & 98\% & 56.60 & 102\% & \multirow{2}{*}{0.004} \\
 & Overlap & Mis & 55.00 & 55.44 & 101\% & 56.05 & 102\% & 52.38 & 95\% & 55.51 & 101\% &  \\ \cline{2-13} 
 & Spelling & Mat & 57.55 & 55.41 & 96\% & 55.80 & 97\% & 55.20 & 96\% & 55.34 & 96\% & \multirow{2}{*}{0.004} \\
 & Error & Mis & 55.53 & 56.01 & 101\% & 55.31 & 100\% & 55.36 & 100\% & 54.95 & 99\% &  \\ \hline
 \hline 
\multirow{10}{*}{DGA} & \multirow{2}{*}{Antonymy} & Mat & 15.02 & 14.32 & 95\% & 15.67 & 104\% & 14.02 & 93\% & 13.22 & 88\% & \multirow{2}{*}{0.073} \\
 &  & Mis & 16.12 & 14.45 & 90\% & 16.38 & 102\% & 13.98 & 87\% & 13.13 & 81\% &  \\ \cline{2-13} 
 & Length & Mat & 45.21 & 45.36 & 100\% & 44.27 & 98\% & 43.65 & 97\% & 42.19 & 93\% & \multirow{2}{*}{0.009} \\
 & Mismatch & Mis & 43.97 & 45.21 & 103\% & 44.01 & 100\% & 42.16 & 96\% & 42.65 & 97\% &  \\ \cline{2-13} 
 & \multirow{2}{*}{Negation} & Mat & 44.91 & 43.28 & 96\% & 44.02 & 98\% & 43.99 & 98\% & 40.18 & 89\% & \multirow{2}{*}{0.025} \\
 &  & Mis & 43.99 & 43.17 & 98\% & 43.87 & 100\% & 43.23 & 98\% & 39.09 & 89\% &  \\ \cline{2-13} 
 & Word & Mat & 52.39 & 50.56 & 97\% & 51.34 & 98\% & 52.42 & 100\% & 49.56 & 95\% & \multirow{2}{*}{0.005} \\
 & Overlap & Mis & 52.06 & 50.87 & 98\% & 51.65 & 99\% & 52.12 & 100\% & 50.23 & 96\% &  \\ \cline{2-13} 
 & Spelling & Mat & 59.37 & 57.36 & 97\% & 58.11 & 98\% & 58.05 & 98\% & 56.36 & 95\% & \multirow{2}{*}{0.011} \\
 & Error & Mis & 60.01 & 58.02 & 97\% & 58.92 & 98\% & 57.34 & 96\% & 55.67 & 93\% &  \\ \hline
 \hline 
\multirow{10}{*}{ESIM} & \multirow{2}{*}{Antonymy} & Mat & 16.91 & 15.88 & 94\% & 16.07 & 95\% & 16.98 & 100\% & 13.23 & 78\% & \multirow{2}{*}{0.079} \\
 &  & Mis & 16.7 & 15.29 & 92\% & 16.22 & 97\% & 15.87 & 95\% & 14.03 & 84\% &  \\ \cline{2-13} 
 & Length & Mat & 44.98 & 40.66 & 90\% & 42.91 & 95\% & 45.43 & 101\% & 41.98 & 93\% & \multirow{2}{*}{0.016} \\
 & Mismatch & Mis & 44.88 & 39.81 & 89\% & 41.99 & 94\% & 44.76 & 100\% & 41.73 & 93\% &  \\ \cline{2-13} 
 & \multirow{2}{*}{Negation} & Mat & 45.16 & 43.18 & 96\% & 44.67 & 99\% & 45.02 & 100\% & 43.01 & 95\% & \multirow{2}{*}{0.005} \\
 &  & Mis & 45.27 & 43.29 & 96\% & 44.55 & 98\% & 45.00 & 99\% & 43.24 & 96\% &  \\ \cline{2-13} 
 & Word & Mat & 57.41 & 46.34 & 81\% & 44.35 & 77\% & 43.21 & 75\% & 37.73 & 66\% & \multirow{2}{*}{0.470} \\
 & Overlap & Mis & 58.27 & 46.20 & 79\% & 45.53 & 78\% & 43.40 & 74\% & 37.52 & 64\% &  \\ \cline{2-13} 
 & Spelling & Mat & 57.09 & 50.66 & 89\% & 48.75 & 85\% & 43.65 & 76\% & 37.40 & 66\% & \multirow{2}{*}{0.370} \\
 & Error & Mis & 56.48 & 50.98 & 90\% & 49.42 & 88\% & 44.31 & 78\% & 37.54 & 66\% &  \\ \hline
 \hline 
\multirow{10}{*}{KIM} & \multirow{2}{*}{Antonymy} & Mat & 83.15 & 78.41 & 94\% & 76.69 & 92\% & 79.15 & 95\% & 76.49 & 92\% & \multirow{2}{*}{0.025} \\
 &  & Mis & 83.04 & 78.43 & 94\% & 80.04 & 96\% & 81.08 & 98\% & 75.26 & 91\% &  \\ \cline{2-13} 
 & Length & Mat & 47.27 & 46.89 & 99\% & 48.80 & 103\% & 48.23 & 102\% & 47.75 & 101\% & \multirow{2}{*}{0.003} \\
 & Mismatch & Mis & 48.88 & 46.47 & 95\% & 48.90 & 100\% & 49.15 & 101\% & 46.96 & 96\% &  \\ \cline{2-13} 
 & \multirow{2}{*}{Negation} & Mat & 51.38 & 47.36 & 92\% & 40.91 & 80\% & 43.97 & 86\% & 38.86 & 76\% & \multirow{2}{*}{0.281} \\
 &  & Mis & 53.04 & 48.18 & 91\% & 40.35 & 76\% & 45.51 & 86\% & 37.85 & 71\% &  \\ \cline{2-13} 
 & Word & Mat & 54.18 & 53.12 & 98\% & 54.32 & 100\% & 53.67 & 99\% & 52.87 & 98\% & \multirow{2}{*}{0.005} \\
 & Overlap & Mis & 54.57 & 53.27 & 98\% & 53.82 & 99\% & 52.55 & 96\% & 51.66 & 95\% &  \\ \cline{2-13} 
 & Spelling & Mat & 60.19 & 59.16 & 98\% & 58.17 & 97\% & 59.98 & 100\% & 57.39 & 95\% & \multirow{2}{*}{0.010} \\
 & Error & Mis & 61.32 & 60.36 & 98\% & 59.21 & 97\% & 59.65 & 97\% & 56.87 & 93\% &  \\ \hline
 \hline 
\multirow{10}{*}{ADV} & \multirow{2}{*}{Antonymy} & Mat & 33.63 & 27.99 & 83\% & 25.82 & 77\% & 20.69 & 62\% & 21.46 & 64\% & \multirow{2}{*}{0.613} \\
 &  & Mis & 30.22 & 29.76 & 98\% & 21.91 & 73\% & 20.88 & 69\% & 20.24 & 67\% &  \\ \cline{2-13} 
 & Length & Mat & 36.23 & 35.98 & 99\% & 36.98 & 102\% & 34.12 & 94\% & 33.23 & 92\% & \multirow{2}{*}{0.022} \\
 & Mismatch & Mis & 36.47 & 35.78 & 98\% & 37.04 & 102\% & 34.23 & 94\% & 33.45 & 92\% &  \\ \cline{2-13} 
 & \multirow{2}{*}{Negation} & Mat & 40.18 & 39.23 & 98\% & 37.38 & 93\% & 36.99 & 92\% & 34.19 & 85\% & \multirow{2}{*}{0.080} \\
 &  & Mis & 40.86 & 39.48 & 97\% & 37.12 & 91\% & 36.78 & 90\% & 34.02 & 83\% &  \\ \cline{2-13} 
 & Word & Mat & 42.19 & 40.43 & 96\% & 39.67 & 94\% & 38.76 & 92\% & 36.43 & 86\% & \multirow{2}{*}{0.066} \\
 & Overlap & Mis & 41.98 & 40.43 & 96\% & 36.97 & 88\% & 38.76 & 92\% & 36.43 & 87\% &  \\ \cline{2-13} 
 & Spelling & Mat & 37.89 & 35.17 & 93\% & 34.88 & 92\% & 34.12 & 90\% & 31.29 & 83\% & \multirow{2}{*}{0.097} \\
 & Error & Mis & 37.63 & 35.26 & 94\% & 34.14 & 91\% & 34.09 & 91\% & 30.76 & 82\% &  \\ \hline
 \hline 
\end{tabular}
\end{table*}

\begin{table*}[]
\centering
\caption{Transfer test results for other downstream tasks when different percentage of training data are swapped.}
\label{tab:result4}
\begin{tabular}{ccccccccc}
\hline
 & MR & CR & MPQA & SUBJ & TREC & SICK-E & SICK-R & MRPC \\ \hline
0\% & 76.92 & 78.15 & 87.64 & 90.79 & 83.6 & 82.3 & 0.859 & 74.09/\textbf{82.45} \\
25\% & 76.26 & 79.13 & \textbf{87.72} & 90.92 & 84.6 & 82.87 & 0.856 & 73.1/81.97 \\
50\% & \textbf{77.3} & \textbf{79.84} & 87.5 & 90.28 & 79.8 & 83.17 & \textbf{0.862} & \textbf{75.25}/82.26 \\
75\% & 76.42 & 79.66 & 87.71 & \textbf{90.99} & 80.4 & \textbf{83.24} & 0.859 & 73.8/82.32 \\
100\% & 75.14 & 77.43 & 87.5 & 90.81 & \textbf{85} & 83.09 & 0.855 & 73.22/82.01 \\ \hline
\end{tabular}
\end{table*}

To test the robustness of an NLI model, we need an evaluation metric that evaluates the model's performance at the semantic level, independent of words' distributions. We consider the recently proposed stress test \citep{naik2018stress}. It is an evaluation method that helps to examine whether the models can predict at the semantic level. They created a test set that is constructed following a variety of different rules, including competence test set (antonym, numerical reasoning), distraction test set (with three strategies: word overlap, negation, and length mismatch), and also noise test set. There are six evaluation criteria altogether. %In competence test, they mainly test the method's ability in reasoning about quantities and antonyms. In the distraction test, they mainly test whether the model is predicting through simple statistics (\textit{i.e.} confounding factors introduced by workers' word choices). In the noise test, they check whether the model is robust to noises such as spelling errors. 

% With the stress test, we can have a better understanding how swapping the word fragments will help in mitigating the confounding factors. To have a more thorough understanding, we test over a spectrum of different percentages of the sampled to be swapped, ranging from 0\% to 100\%. 
Following \citep{naik2018stress}, we trained the NLI models discussed in the previous section on MultiNLI data set \citep{williams2017broad} and tested on the genre-matched and mismatched cases separately. We trained the models with five different cases (\textit{i.e.} when 0\%/25\%/50\%/75\%/100\% of training pairs are swapped) and reported the testing scores in Table~\ref{tab:result3}. S$\star$ reports scores with different training data, and R$\star$ reports the relative changes of the corresponding case (ratio of S$\star$ over S0\%). We also calculated the deviation for each stress test, which indicates the robustness of the model to confounding factors (the smaller, the better). The deviation is calculated as the sum of the squared error between R$\star$ and 100\% and is reported in the last column.  

By looking into the last column of Table~\ref{tab:result3}, we can see that different methods showed a different level of robustness. Overall, DGA is least affected by the spurious signals, which is consistent with our results in the previous section. For other methods, InferSent and ADV show robustness except on Antonymy test, KIM shows robustness except on Negation test (which is also consistent with the findings in the previous section), and ESIM shows robustness except in Word Overlap and Spelling Error case. 

We do not expect that swapping the training data pairs will help the model in dealing with spelling errors so that all the models should be robust in the Spelling Error test case. However, interestingly, ESIM shows a surprising drop of performance in Spelling Error test, and the magnitude of performance drop seems to correlate with the amount of data swapped. Other drops that are correlated with the amount of data swapped include the Word Overlap test for ESIM and Antonymy test for ADV. 

Also, we notice KIM leads in test scores and the Antonymy test for KIM is impressively higher than any other competitors. This result is probably due to that KIM utilizes an external knowledge base that includes antonymy information. 

%Interestingly, we notice that the swapping training procedure can help improve the test performance when a reasonable amount of text fragments are swapped (\textit{e.g.} 50\% to 75\%). It is also exciting to see that swapping training contribute significantly in the Antonymy case. In other cases, although the improvement is marginal, it is still exciting to see that we can almost guarantee to improve the performance when we randomly swap 50\%-75\% of the training fragments. 

%Another interesting phenomenon we observe is that swapping training does not help in the Spelling error test at all. This may look discouraging at first, but as we have discussed, swapping training is introduced to mitigate the confounding factor problems mentioned previously, it is not supposed to improve performance in the spelling error test. 

\section{Swapping Training for NLP Transfer Tasks}
\label{sec:transfer}
% As our swapping evaluation metrics have shown, many recently proposed NLI methods achieved high predictive performances by learning through the workers' word choices as confounding factors. These results unfortunately leads to a conclusion that the community need to develop newer NLI data sets with more rigorous control on how the data should be constructed. However, constructing new data sets takes efforts and time. Therefore, in this paper, we continue to ask and answer an interesting question: what is a simple and efficient way to take advantages of current data sets and NLI models to serve the NLP community before newer data set is constructed. 

% To answer this question, we proceed to propose a swapping training techniques that we randomly swap the \textit{premises} and \textit{hypothesis} during training phases, so that we can mitigate the confounding factor problem. However, we should notice that, such swapping can only occur for \textit{contradiction} pairs and \textit{neutral} pairs because, as we have shown, the labels of \textit{entailment} pairs do not preserve for swapping operation. 
Finally, we study how the swapping operation will help mitigate the confounding factor problem by training a general-purpose sentence embedding that can capture the generic information for other transfer NLP tasks. Following \citep{conneau2017supervised}, we re-train the InferSent model with the swapping operation and compare the performance when different percentages of sentence pairs are swapped. 

We evaluated the sentence embedding with a set of different transfer NLP tasks \citep{conneau2017supervised}, including several basic classification tasks such as sentiment analysis (MR, SST), question-type (TREC), product reviews (CR), subjectivity/objectivity (SUBJ) and opinion polarity (MPQA), paraphrase identification (MRPC), and entailment (SICK-E) and semantic relatedness (SICK-R) from SICK data set \citep{marelli2014sick}. 

These transfer tasks are evaluated with the standard metrics applied on these tasks. The MR, SST, TREC, CR, SUBJ and MPQA are evaluated with accuracy \citep{conneau2017supervised}. The MRPC is evaluated with both accuracy and F1 \citep{subramanian2018learning}. The SICK-E and SICK-R are evaluated with Pearson correlation \citep{tai2015improved}.  

We report the results in Table~\ref{tab:result4}. As we can see, although the improvement seems marginal, the swapping operation helps improve these transfer tasks because the model is often evaluated as the best when 25\%-75\% of the sentences are swapped. 

\section{Discussion}
One may argue that the \textit{neutral} pairs may not remain \textit{neutral} after the two text fragments swapped, especially when the \textit{hypothesis} is more specific than \textit{premise}. For example, with a \textit{premise} ``I bought books'', several \textit{hypothesis} can be generated to change the NLI label, such as ``I bought 5 books'', ``I bought books on history'', or ``I bought books in the bookstore near campus''. However, through this paper, we assume the label preserves in most cases. This assumption is verified empirically by results in Table~\ref{tab:result1}: the performance differences due to swapping are very similar in \textit{contradiction} case and \textit{neutral} case in most of the methods tested, leading to a conjecture that the label preserving property of swapping operation will be similar for \textit{contradiction} and \textit{neutral} in the data set.   

\section{Related Work} 
\label{sec:related}
%Before we conclude the paper, we present a summary of related works. We will first discuss the recent computational advances in NLI tasks. Then we will discuss the confounding factor problems in other tasks.  

\paragraph{Computational Methods for NLI}
Ever since the introduction of recent large scale NLI data set \citep{bowman2015large}, many recent advanced computational models have been proposed, and a majority of these are LSTM methods or Bidirectional LSTM with extensions \citep{rocktaschel2015reasoning,wang2015learning,bowman2016fast,liu2016deep,vendrov2015order,liu2016learning,liu2016modelling,cheng2016long,sha2016reading,munkhdalai2017neural,munkhdalai2017neural,nie2017shortcut,choi2018learning,peters2018deep}.

Here we offer a brief overview of the most recent methods applied to NLI tasks: \citet{tay2017compare} proposed CAFE (Compare-propagate Alignment- Factorized Encoders). The essential component of CAFE is a compare-propagate architecture which first compares the two text fragments and then propagate the aligned features to upper layers for representation learning. 
\citet{shen2018reinforced} presented reinforced self-attention (ReSA), which aims to combine the benefit of soft attention and a newly proposed hard attention mechanism called reinforced sequence sampling (RSS). They further plugged this ReSA onto a source2token self-attention model and applied to NLI tasks. 
\citet{kim2018semantic} proposed a a densely-connected co-attentive recurrent neural network, whose essential idea is that the recurrent component uses the concatenated information from any layer to all the subsequent layers. They also used an autoencoder after dense concatenation to reduce the problem of ever-increasing sizes of representations. 
\citet{chen2018enhancing} introduced a vector-based multi-head attention as a generalized pooling method a weighted summation of hidden vectors to enhance sentence embedding. This method is then built on a bidirectional LSTM and applied to NLI tasks. \citet{tan2018multiway} proposed a multiway attention network, which combines the information form four attention word-matching functions defined by four mathematical operations to build up the representation. They further built the proposed method on Gated Recurrent Unit (GRU) and applied to NLI tasks. 
\citet{liu2018stochastic} introduced a stochastic answer network (SAN) for multi-step inference strategies for NLI. Different from conventional methods that directly predict given input sentence pairs, the SAN maintains a state and iteratively refines the predictions. 

\paragraph{Confounding factor problems in other tasks and corresponding solutions}
In a broader domain other than NLI, many other machine learning methods noticed the problems introduced via confounding factors, which lead the machine learning methods yield a higher predictive performance than what a model can achieve in a confounder-free setting. 
For example, \citet{wang2016select} noticed that in video sentiment analysis, a random split of data into training set and testing set will typically yield much higher testing performance than split the data to make sure the data samples in testing set and training set will never come from the same individual (though the samples are different) because some recognizable features of the individual serve as confounding factors. Further, they proposed a select-additive learning method to mitigate the problem. 
\citet{goyal2017making} noticed that in visual question answering, the model would generate the answers for the images mainly based on the distribution of words of the training data, instead of the association between the images and the sentences in the training data. They further reorganize the data to balance the distributions to avoid such problems. 
The confounding-factor problem needs more attention in biomedical applications where usually all the side information such as gender and age can serve as confounding factors \citep{yue2018deep}. A recent paper empirically discusses the challenges \citep{zech2018confounding}, and a recent solution \citep{wang2018removing} mitigated the confounding factor challenge. 

\section{Conclusion}
In this paper, we first discussed the existence of confounding factors of words' distribution of \textit{hypothesis} in NLI data set due to the construction of data sets. Then with simple propositional logic rules, we presented a simple sanity check for the computational methods for natural language inference. Our argument is that for an evaluation data set where the \textit{premise} and \textit{hypothesis} are swapped, \textbf{if a model can truly predict inference relationship between pairs of text fragments, it should report comparable accuracy between the original test set and swapped test set for \textit{contradiction} pairs and neural pairs, and lower accuracy in swapped test set for \textit{entailment} pairs.} We applied our swapping evaluation towards several recently proposed models, and the results revealed some interesting properties of these methods. 

Further, we proposed to train the NLI models with a sequence of different training data sets defined by the percentages of different sentence pairs swapped in the training set. We then test these models with the stress test and investigate how the evaluated performance fluctuates for each model. We used ``deviation'' to measure the fluctuation, which is an indication of the robustness of the models to confounding factors of words' distribution. The robustness testing offered some other understandings of these NLI models. Overall, our swapping testing and robustness testing indicate that DGA \citep{chen2017recurrent} and KIM \citep{chen2018neural} are powerful at the semantic level and robust to the confounding factors. ADV \citep{minervini2018adve} is also a promising method, but some more detailed studies are recommended to be conducted for the interesting properties our metrics revealed. 

Finally, we also tested to see how swapping operation can help mitigate the confounding factor problem by applying the trained sentence embedding to other NLP transfer tasks. We achieved a higher performance on most transfer tasks when 25\%-75\% of the sentence pairs are swapped.

\label{sec:con}

%\subsubsection{Acknowledgments.}

% \subsubsection{Appendices.}
% Any appendices follow the acknowledgments, if included, or after the main body of text if no acknowledgments appear. 

\bibliography{ref}
\bibliographystyle{aaai}

\newpage 

\beginsupplement

\section*{Appendix}

\section{Error Analysis of Table 1}
While Table~\ref{tab:result1} has revealed several interesting properties of these NLI methods, an important question left is whether these drops of performance are because of the internal properties of the data sets, or more model-specific. 

To answer this question, we collected the 400 sentence pairs that are most frequently mis-classified along all the epoches for each of those five models. We then investigate the overlaps of these samples and report the result in Table~\ref{tab:append:1}

\begin{table}[!htbp]
\caption{The overlaps of top 400 most frequently misclassification examples of each method once swapped}
\label{tab:append:1}
\begin{tabular}{|c|ccccc|}
\hline
 & InferSent & DGA & ESIM & KIM & ADV \\ \hline
InferSent & - & 0 & 0 & 5 & 0 \\
DGA & 0 & - & 0 & 0 & 0 \\
ESIM & 0 & 0 & - & 0 & 5 \\
KIM & 5 & 0 & 0 & - & 7 \\
ADV & 0 & 0 & 5 & 7 & - \\ \hline
\end{tabular}
\end{table}

As Table~\ref{tab:append:1} shows, out of these 400 samples each, there are only 5 overlaps between InferSent and Kim, 5 overlaps between ADV and ESIM, and 7 overlaps between ADV and KIM. Therefore, the misclassification is more model specific. 

We also calculate the distribution of six different types of misclassifications (E$\rightarrow$N, E$\rightarrow$C, N$\rightarrow$E, N$\rightarrow$C, C$\rightarrow$E, C$\rightarrow$N, where we denote the misclassification as ``labels''$\rightarrow$``prediction''), and we notice that models showed a slight deviation from the averaged distribution overall. Particularly, ADV results in the least fraction in C$\rightarrow$N out of these five methods, DGA results in the least fraction in E$\rightarrow$N cases, and InferSent results in the least fraction in N$\rightarrow$C, as shown by Table~\ref{tab:append:2}.  
 
\begin{table}[!htbp]
\centering
\caption{The fractions of mis-classification examples for each model}
\label{tab:append:2}
\begin{tabular}{cccccc}
\hline
 & InferSent & DGA & ESIM & KIM & ADV \\ \hline
E$\rightarrow$N & 0.1175 & 0.0850 & 0.1125 & 0.1475 & 0.1075 \\
E$\rightarrow$C & 0.0325 & 0.0250 & 0.0425 & 0.0350 & 0.0275 \\
N$\rightarrow$E & 0.2050 & 0.2000 & 0.2250 & 0.2275 & 0.2225 \\
N$\rightarrow$C & 0.1800 & 0.2475 & 0.1825 & 0.2375 & 0.2050 \\
C$\rightarrow$E & 0.0325 & 0.0225 & 0.0375 & 0.0275 & 0.0225 \\
C$\rightarrow$N & 0.4325 & 0.4200 & 0.4000 & 0.3250 & 0.4150 \\ \hline
\end{tabular}
\end{table}

\end{document}